\documentclass[pdflatex,sn-mathphys-ay]{sn-jnl}% Math and Physical Sciences Author Year Reference Style
\usepackage{graphicx}%
\usepackage{multirow}%
\usepackage{amsmath,amssymb,amsfonts}%
\usepackage{amsthm}%
\usepackage{mathrsfs}%
\usepackage[title]{appendix}%
\usepackage{xcolor}%
\usepackage{textcomp}%
\usepackage{manyfoot}%
\usepackage{booktabs}%
\usepackage{algorithm}%
\usepackage{algorithmicx}%
\usepackage{algpseudocode}%
\usepackage{listings}%
\newcommand{\smallorcid}[1]{%
  \href{https://orcid.org/#1}{%
    \raisebox{-0.1\height}{\includegraphics[height=0.85em]{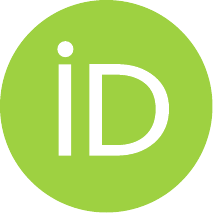}}}}

\theoremstyle{thmstyleone}%
\theoremstyle{thmstyletwo}%

\theoremstyle{thmstylethree}%

\begin{document}

\title[Physics-Informed Self-Supervised Learning for Joint Wire Calibration and Interaction Position Reconstruction in Multi-Wire Parallel Plate Avalanche Counters
]{Physics-Informed Self-Supervised Learning for Joint Wire Calibration and Interaction Position Reconstruction in Multi-Wire Parallel Plate Avalanche Counters
}


\author[1]{\fnm{Antoine} \sur{Lemasson}\,\smallorcid{0000-0002-9434-8520}}\email{lemasson@ganil.fr}

\author[1]{\fnm{Maurycy} \sur{Rejmund}\,\smallorcid{0009-0009-8626-8756}}\email{mrejmund@ganil.fr}
\affil[1]{GANIL, CEA/DRF - CNRS/IN2P3, Bd Henri Becquerel, BP 55027, F-14076 Caen Cedex 5, France}

\abstract{
Scientific instruments require accurate calibration to convert detector signals into 
reliable physical observables. Conventional calibration procedures typically rely on 
dedicated calibration measurements, analytical response models, or labelled reference 
data, limiting their ability to adapt to changing operating conditions and detector aging. 

We present a physics-informed self-supervised learning framework that jointly performs 
wire calibration and interaction position reconstruction in Multi-Wire Parallel Plate Avalanche 
Counters (MWPPACs) without requiring labelled position measurements or dedicated calibration 
runs. The proposed method formulates detector calibration as a latent optimization problem in 
which global wire gains and event-wise interaction positions are estimated simultaneously using 
supervision derived exclusively from detector geometry and charge-energy consistency constraints. 
A detector-independent neural network reconstructs sub-wire interaction positions from local 
charge distributions, eliminating the need to assume analytical induction profiles by learning the 
detector response directly from experimental data. The resulting end-to-end differentiable framework 
enables continuous detector self-calibration while improving the uniformity and accuracy of 
position reconstruction. Experimental evaluation on the entrance MWPPAC tracking detectors 
of the VAMOS++ magnetic spectrometer demonstrates stable convergence, improved spatial 
homogeneity, and enhanced position resolution. 

Beyond the specific detector studied, the proposed methodology establishes a general framework for 
physics-informed self-supervised calibration of scientific instruments and represents a significant step 
toward autonomous intelligent instrumentation capable of continuous adaptation during operation. 
In this paradigm, detector calibration is no longer a prerequisite for an experiment but an integral part 
of the measurement process itself.}


\maketitle

\section{Introduction}

Scientific instruments rely on accurate calibration to convert raw sensor responses into
physically meaningful observables. Traditionally, detector calibration is performed
through dedicated calibration measurements, analytical parameter estimation,
or supervised regression using labelled reference data.
Although these approaches have proven effective for many instrumentation systems,
they present important practical limitations. Calibration procedures are often
experiment specific, require interruptions of normal operation, depend on dedicated
reference measurements, and may become invalid as detector characteristics evolve
due to changes in operating conditions, electronic response, mechanical deformation,
or detector aging. As a consequence, maintaining optimal detector performance
throughout extended experimental campaigns remains a significant challenge.

Recent advances in machine learning have enabled data-driven approaches to
scientific instrumentation, allowing complex nonlinear detector responses to be
modeled directly from experimental data.
Most existing methods, however, remain fundamentally supervised and therefore
depend on the availability of accurately labelled calibration data.
For many scientific instruments, obtaining such labels is difficult,
expensive, or even impossible.
This has motivated increasing interest in self-supervised learning, where
the supervisory signal is derived from intrinsic physical relationships
rather than externally provided annotations.

In our recent work~(\citet{Rejmund2026arxiv}),
we introduced a general framework for the self-supervised calibration of
scientific instruments using physical consistency constraints.
Instead of relying on labelled calibration samples,
unknown detector calibration parameters were optimized jointly with the desired
physical observables by minimizing objective functions derived exclusively from
known physical laws.
The resulting formulation naturally integrates domain knowledge into the learning
process while enabling continuous calibration during normal detector operation.

In the present work, we extend this paradigm to one of the most challenging
subsystems of the VAMOS++ magnetic spectrometer~(\citet{Pullanhiotan2008, Rejmund2011}), 
namely the Dual Position Sensitive MWPPAC tracking detector~(\citet{Vandebrouck2016}).
Unlike conventional calibration procedures, which typically estimate detector
gains independently from position reconstruction and often assume an analytical
shape for the induced charge distribution, the proposed framework treats both
problems simultaneously.
Wire calibration and event-wise interaction position reconstruction are formulated
as a single self-supervised optimization problem in which neither the wire gains
nor the interaction positions are directly observed.
Instead, both quantities are inferred jointly by minimizing losses derived from
detector geometry and charge-energy consistency.

A key feature of the proposed approach is the separation between global and local
latent variables.
Wire calibration coefficients represent global detector parameters that are shared
by all events, whereas interaction positions constitute event-specific latent
variables.
These two sets of unknown quantities are coupled through differentiable physical
constraints, allowing both detector calibration and position reconstruction to be
optimized simultaneously using gradient-based learning without requiring labelled
position measurements or dedicated calibration runs.

Another important aspect of the proposed framework is that the position
reconstruction network is independent of detector geometry.
Instead of learning absolute detector coordinates, the network estimates only the
sub-wire interaction offset from the calibrated charge distribution observed on a
small neighborhood of adjacent wires.
Consequently, the learned mapping depends solely on the local induction profile
and can therefore be transferred to detectors with different dimensions or numbers
of sensing wires.
Moreover, no analytical assumption regarding the induced charge distribution is
required, allowing the underlying detector induction profile to emerge directly from 
experimental data without prescribing any analytical functional form.

The proposed methodology provides several practical advantages.
It enables continuous detector self-calibration during normal operation,
eliminates the need for dedicated calibration measurements,
avoids assumptions regarding analytical charge distributions,
and naturally adapts to slow detector evolution arising from ageing,
electronic drifts, or changing experimental conditions.
Although demonstrated here using the MWPPAC tracking detectors of VAMOS++,
the proposed framework is applicable to a broad class of position-sensitive
detectors whose operation is governed by known physical consistency relations.

The principal contributions of this work are summarized as follows:

\begin{itemize}
\item formulation of MWPPAC wire calibration and interaction position reconstruction 
as a unified self-supervised latent optimization problem;

\item development of a physics-informed learning framework that jointly estimates global 
wire calibration coefficients and event-wise interaction positions;

\item introduction of a detector-independent position reconstruction network that learns 
the local induction profile directly from experimental data without assuming an analytical charge distribution;

\item demonstration that detector geometry and charge-energy consistency provide sufficient 
supervisory information to achieve accurate calibration and position reconstruction without labelled data.
\end{itemize}

\section{Related Work}

Machine learning has become an increasingly important tool for scientific
instrumentation, enabling nonlinear detector responses to be reconstructed
with significantly higher accuracy than conventional analytical methods.
Within nuclear physics, neural networks have been successfully employed for
particle identification, trajectory reconstruction, detector calibration,
signal denoising, and event classification.

For the VAMOS++ magnetic spectrometer, deep neural networks have previously
been applied to multidimensional trajectory reconstruction, substantially
improving the determination of particle trajectories within the spectrometer~(\citet{Rejmund2025a}).
Subsequent work introduced fractionally labelled learning for the simultaneous
determination of ionic charge states and atomic numbers from ionization chamber
signals, demonstrating that machine learning can efficiently exploit complex
correlations present in detector responses while reducing the dependence on
fully labelled experimental data~(\citet{Rejmund2025b}).

More recently, self-supervised learning has emerged as an attractive alternative
for scientific applications where labelled calibration data are unavailable.
Instead of relying on externally provided targets, these methods derive
supervision from physical laws, conservation principles, or internal consistency
constraints inherent to the measurement process~(\citet{Rejmund2026arxiv}).
Such approaches belong to the broader field of scientific machine learning,
where physical knowledge is incorporated directly into optimization.

The present work builds upon our recently proposed framework for
self-supervised calibration of scientific instruments using physical
consistency constraints.
While the previous formulation addressed the calibration of ionization chamber
responses for ionic charge-state reconstruction, the problem considered here is
fundamentally different.
The unknown detector calibration now consists of individual wire gains,
whereas the desired observables correspond to interaction positions that are
themselves unknown.
Consequently, calibration and reconstruction become mutually dependent and
must be optimized simultaneously.

Existing position reconstruction methods for multi-wire detectors generally
assume a predefined analytical model of the induced charge distribution,
such as Gaussian or related parametric profiles, before estimating the
interaction position through analytical fitting.
The validity of these assumptions depends on detector geometry, operating
conditions, and electronic response, and may degrade over time.
In contrast, the proposed method makes no explicit assumption regarding the
functional form of the induction profile.
Instead, the relationship between neighboring wire charges and sub-wire
interaction position is learned directly from experimental observations while
remaining constrained by detector geometry and charge-energy consistency.

To the best of our knowledge, this is the first self-supervised framework that
jointly optimizes wire calibration and interaction position reconstruction in
MWPPAC tracking detectors using only physically derived consistency constraints,
without requiring labelled interaction positions or dedicated calibration data.

\section{Physics-Informed Self-Supervised Formulation}

The objective of the proposed framework is to simultaneously estimate the
calibration of individual MWPPAC sensing wires and reconstruct the interaction
position of every detected particle without requiring labelled position
measurements or dedicated calibration data.
Unlike conventional supervised learning, neither the calibration coefficients
nor the interaction positions are known during training.
Both quantities therefore constitute latent variables that must be inferred
jointly from the measured detector responses under physically motivated
consistency constraints.

Throughout this work, the dual MWPPAC detector is regarded as a redundant
measurement system.
Each detector independently measures the horizontal and vertical interaction
coordinates of the same charged particle while the known detector geometry
provides deterministic relationships between the corresponding interaction
positions.
These geometric relationships, together with charge-energy consistency constraints,
provide the supervisory signal required for self-supervised optimization.

\begin{figure*}[htbp]
  \centering
  \includegraphics[width=1\textwidth]{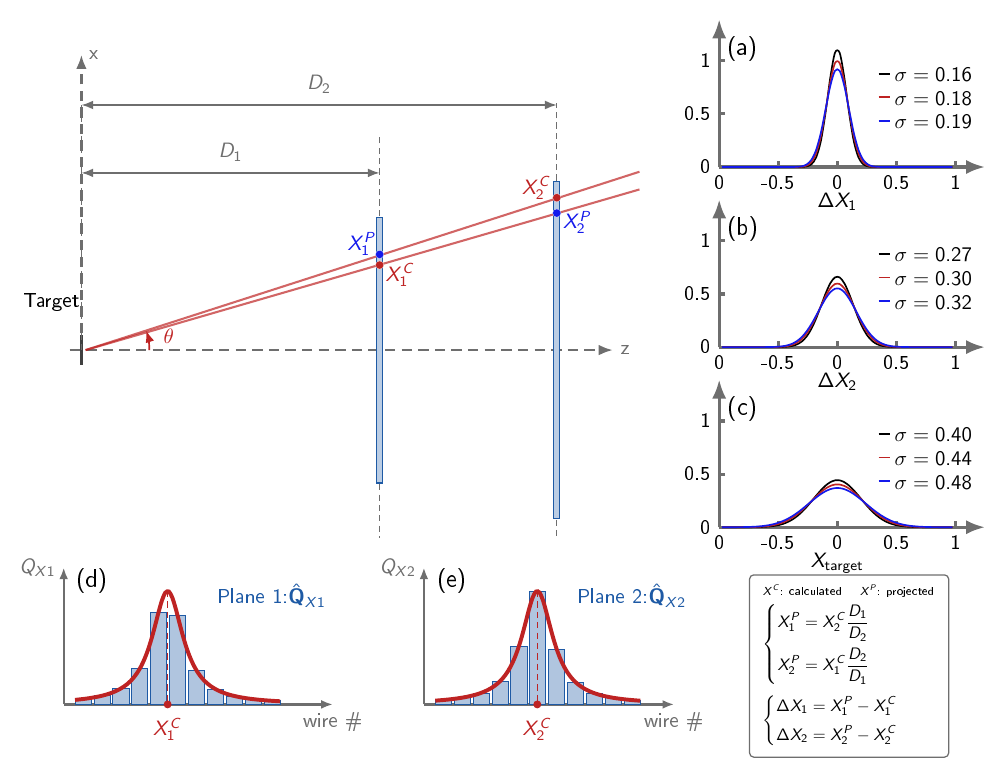}
  \caption{{\bf Detector Geometry:} 
    A particle emitted from the target traverses the two position-sensitive MWPC planes 
    at distances $D_1$ and $D_2$. For each plane, the interaction position $X_k^{C}$ is 
    determined from the charge distribution $\hat{Q}_k$ as illustrated panels (d) and (f). 
    The geometrically projected interaction position $X_k^{P}$ is derived from the interaction 
    position on the complementary 
    plane. The residuals $\Delta X_k = X_k^{P} - X_k^{C}$, which ideally should be zero, 
    contribute to the training target.  The right-side insets 
    depict the simulated distributions of residuals $\Delta X_1$ (a) and $\Delta X_2$ (b) 
    and the reconstructed distribution of the beam on the target $X_{\rm target}$ (c), assuming  
    a true beam width $\sigma_{X_{\rm beam}} = 0.40$~mm under 
    different conditions:
    (black) detector resolution 
    $\sigma_{X_{\rm det}} = 0$~mm,
    (red) $\sigma_{X_{\rm det}} = 0.06$~mm, 
    and
    (blue) $\sigma_{X_{\rm det}} = 0.09$~mm. The resulting widths ($\sigma$) are indicated in mm.
    }
  \label{fig:DetectorGeometry}
\end{figure*}

\subsection{Detector Geometry}

The tracking detector consists of two position-sensitive MWPPAC detectors
located upstream of the VAMOS++ magnetic spectrometer.
Each detector independently measures the horizontal ($X$) and vertical ($Y$)
interaction coordinates together with the timing information required for
trajectory reconstruction. The detector geometry is depicted in Figure~\ref{fig:DetectorGeometry}.

The first detector is located at a distance $D_1$ from the reaction target,
whereas the second detector is positioned at distance $D_2$.
Assuming that the outgoing reaction products originate from the target region
and propagate along straight trajectories over the  distance
between both MWPPAC detectors, the interaction position measured on one detector
uniquely predicts the corresponding interaction position on the other detector.

Denoting the reconstructed interaction positions by
$X_1^C$, $X_2^C$, $Y_1^C$, and $Y_2^C$,
the corresponding predicted positions are
\begin{equation}
X_2^P=X_1^C\frac{D_2}{D_1},
\label{eq:X2P}
\end{equation}

\begin{equation}
X_1^P=X_2^C\frac{D_1}{D_2},
\label{eq:X1P}
\end{equation}
with analogous expressions for the vertical coordinate.

Perfect detector calibration and position reconstruction imply that the
predicted and reconstructed interaction positions coincide.
Therefore, the geometric residuals
\begin{equation}
\Delta X_1=X_1^P-X_1^C,
\label{eq:DX1}
\end{equation}
\begin{equation}
\Delta X_2=X_2^P-X_2^C,
\label{eq:DX2}
\end{equation}
and similarly $\Delta Y_1$ and $\Delta Y_2$,
should ideally vanish for every event.

\subsection{Latent Variables}

The proposed framework jointly estimates two fundamentally different categories
of unknown quantities.

The first category comprises the calibration coefficients of every sensing wire.
These coefficients represent global detector parameters that remain constant
throughout a given calibration period and are shared by all recorded events.
Each detector plane possesses its own independent calibration vector owing to
differences in detector construction, electronic response, operating voltages,
mechanical tolerances, and detector aging.
No assumptions regarding spatial smoothness or correlations between neighboring
wires are imposed, allowing every sensing wire to evolve independently.

The second category consists of the interaction positions of individual
particles.
Unlike the calibration coefficients, interaction positions are event-specific
latent variables that differ for every detected particle.
The optimization problem therefore simultaneously involves global detector
parameters shared by the entire dataset and local event parameters unique to
every observation.

The interaction position within a detector plane is represented as
\begin{equation}
X_k^C=n_{X_k}^{\mathrm{max}}+\delta x_k,
\label{eq:XKC}
\end{equation}
where $n_{X_k}^{\mathrm{max}}$ denotes the wire collecting the maximum induced
charge and $\delta x_k$ represents the corresponding sub-wire interaction
offset.
An analogous representation is adopted for the vertical detector planes.

Only the local neighborhood of sensing wires surrounding
$n_{X_k}^{\mathrm{max}}$ is used for position reconstruction.
Consequently, the estimation of $\delta x_k$ depends exclusively on the local
charge distribution and remains independent of detector dimensions or absolute
interaction position.

\subsection{Physical Consistency Constraints}

Since neither detector calibration nor interaction positions are directly
observed, the optimization is driven entirely by physical consistency
constraints.

The first constraint originates from detector geometry.
Because both MWPPAC detectors observe the same particle trajectory, the
reconstructed interaction positions must satisfy the geometric relationships
defined by Eqs.~\ref{eq:X2P} and~\ref{eq:X1P}.

The second constraint exploits the proportionality between the total induced
charge collected by each detector plane and the corresponding energy loss
measured in the ionization chamber.
For every detector plane, the calibrated wire charges satisfy
\begin{equation}
\sum_i Q_i^{\mathrm{cal}}=fE_{\mathrm{IC}},
\end{equation}
where $E_{\mathrm{IC}}$ denotes the ionization chamber segment energy signal and
$f$ is a fixed scaling coefficient selected such that the resulting wire
calibration coefficients remain within a prescribed dynamic range.
Since interaction position depends only on the relative calibration of
neighboring wires, the absolute value of $f$ is not physically relevant.

Together, detector geometry and charge-energy consistency provide sufficient
supervisory information to jointly estimate detector calibration and
interaction positions without requiring labelled experimental data.

\subsection{Joint Optimization Problem}

The objective of the proposed framework is to identify the detector calibration
parameters and event-wise interaction positions that simultaneously satisfy all
physical consistency constraints.

Formally, let
\begin{equation}
\Theta=
\{
\mathcal{C}_{X_1},
\mathcal{C}_{Y_1},
\mathcal{C}_{X_2},
\mathcal{C}_{Y_2},
\mathcal{P}
\}
\end{equation}
denote the complete set of unknown latent variables, where
$\mathcal{C}$ represents the calibration vectors of the individual detector
planes and $\mathcal{P}$ denotes the mapping that reconstructs sub-wire
interaction positions from locally calibrated charge distributions.

The optimization problem can then be expressed as
\begin{equation}
\Theta^\star
=
\arg\min_{\Theta}
\mathcal{L}(\Theta),
\end{equation}
where the objective function $\mathcal{L}$ is constructed exclusively from
physics-derived consistency constraints.
No labelled interaction positions, reference trajectories, or externally
measured calibration coefficients are required during optimization.

In the following section, the proposed neural architecture is introduced as a
differentiable parameterization of the latent variables, enabling efficient
end-to-end optimization using standard gradient-based learning algorithms.

\begin{figure}[htbp]
  \centering
  \includegraphics[width=1\columnwidth]{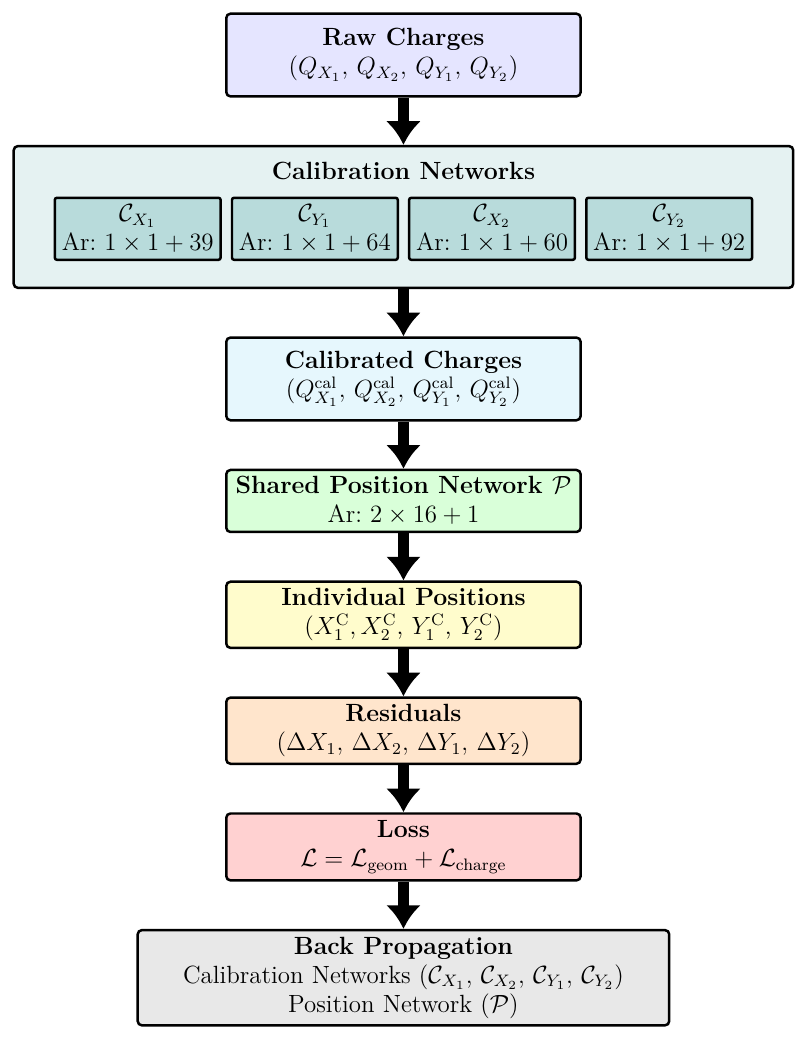}
  \caption{{\bf Network Architecture:} 
    A schematic representation of the network architecture utilized in the current work.
    The architecture of each neural network sub-model is presented in the following format: Ar: $N_l \times N_u + N_{ou}$
    comprising $N_l$ layers and $N_u$ units (neurons) per layer, followed by $N_{ou}$ output units.
    }
  \label{fig:Architecture}
\end{figure}

\section{Joint Neural Parameterization}

The optimization problem formulated in the previous section involves two
distinct categories of latent variables: global detector calibration parameters
shared by all events and local interaction positions that are estimated
independently for every detected particle.
To enable efficient gradient-based optimization, both categories are
parameterized by differentiable neural networks and optimized jointly in an
end-to-end fashion.

The proposed framework consists of two types of neural models.
Four independent calibration subnetworks estimate the gain coefficients of the
individual detector planes, whereas a single shared position reconstruction
network estimates the sub-wire interaction position from locally calibrated
charge distributions.
The complete architecture is illustrated in Fig.~\ref{fig:Architecture}.

\subsection{Calibration Networks}

Each detector plane possesses an independent calibration model responsible for
estimating the gain coefficients of all sensing wires within that plane.
Separate models are employed because every detector plane constitutes an
independent physical system with its own electronic response, operating voltage,
mechanical tolerances, and aging characteristics.
Accordingly, no assumptions regarding shared calibration parameters are imposed
between different detector planes.

For a detector plane containing $M$ sensing wires, the corresponding calibration
network predicts the complete calibration vector
\begin{equation}
\mathcal{C}
=
(c_1,c_2,\ldots,c_M),
\end{equation}
where $c_i$ denotes the multiplicative gain correction associated with the
$i$-th sensing wire.

Unlike conventional neural networks, the calibration coefficients represent
global detector properties rather than event-dependent quantities.
Consequently, the network output should remain independent of the measured
signals from individual events.
To preserve compatibility with standard deep-learning frameworks while allowing
the calibration coefficients to be optimized by back-propagation, the network
receives a dummy scalar input sampled from a uniform distribution,
\begin{equation}
x\sim\mathcal U(0,1000).
\end{equation}

The input carries no physical information and serves exclusively to instantiate
a differentiable computational graph.
Because the desired outputs are constant detector parameters, the calibration
network naturally converges towards an input-independent mapping whose outputs
represent the optimized wire gain coefficients.

Finally, the network outputs are mapped through a sigmoid activation and
linearly rescaled to the admissible calibration interval
\begin{equation}
c_i\in[c_{\min},c_{\max}],
\end{equation}
where in the present work
\[
c_{\min}=0.5,
\qquad
c_{\max}=2.0.
\]

The calibrated wire charges are therefore obtained as
\begin{equation}
Q_i^{\mathrm{cal}}
=
c_iQ_i.
\end{equation}
Since detector pedestals are determined independently during the standard
electronic calibration procedure, only multiplicative gain corrections are
estimated by the proposed framework.

\subsection{Universal Position Reconstruction Network}

In contrast to the calibration networks, the position reconstruction model
operates on an event-by-event basis.
A single neural network is shared by all detector planes, allowing the learned
mapping to remain independent of detector dimensions, wire numbering, and
absolute interaction position.

For every detected event, the sensing wire collecting the largest calibrated
charge is identified.
A local neighborhood containing $N$ adjacent sensing wires centered on this
maximum is then extracted,
\begin{equation}
\hat{Q}
=
(Q_{-k},\ldots,Q_0,\ldots,Q_{+k}),
\label{eq:QVEC}
\end{equation}
where
\[
N=2k+1.
\]
Only odd values of $N$ are considered to preserve symmetry around the central
wire.
Throughout this work, values ranging from $N=3$ to $N=11$ were investigated.

The position network receives only the locally calibrated charge vector
$\hat{Q}$ as input.
Neither detector identity, wire number, nor absolute detector coordinates are
provided.
Instead, the network estimates the sub-wire interaction offset
\[
\delta x\in[-2.0,2.0],
\]
which is subsequently combined with the index of the maximum-charge wire (see Eq.~(\ref{eq:XKC})),

\begin{equation}
X^C=n^{\max}+\delta x.
\end{equation}
An analogous procedure is performed independently for the vertical detector
planes.

Because the network observes only local charge distributions, it effectively
learns the relationship between the induced charge profile and the corresponding
sub-wire interaction position.
Consequently, no analytical assumptions regarding the functional form of the
induction profile are required.
Instead, the induction profile is learned directly from experimental data during
self-supervised optimization.

Moreover, since the mapping depends exclusively on local charge distributions,
the learned model can be directly transferred to detectors possessing different
numbers of sensing wires or different active areas without modification of the
network architecture.

\subsection{Event Processing Pipeline}

For every recorded event, the four detector planes are processed independently
by their respective calibration subnetworks.
The calibrated charge distributions are subsequently used to extract the local
wire neighborhoods centered on the maximum induced charge.

Each extracted neighborhood is passed through the shared position
reconstruction network, producing the corresponding sub-wire interaction
offsets.
These offsets are combined with the indices of the maximum-charge wires to
obtain the reconstructed interaction positions for all detector planes.

Finally, the reconstructed interaction positions and calibrated charge sums are
combined through the physical consistency constraints introduced in the previous
section to evaluate the self-supervised objective function.

\subsection{End-to-End Optimization}

The proposed framework is optimized jointly using standard back-propagation.
Gradients originating from the physics-informed objective function propagate
simultaneously through the position reconstruction network and the calibration
subnetworks.

An important consequence of this joint optimization is the mutual coupling
between detector calibration and position reconstruction.
Improved wire calibration produces more accurate local charge distributions,
which in turn facilitate more precise interaction position estimates.
Conversely, improved position reconstruction yields stronger geometric
consistency, providing more informative supervisory signals for calibration.
Both components therefore evolve simultaneously throughout training until a
globally consistent solution satisfying the physical constraints is obtained.

Unlike conventional sequential calibration procedures, no intermediate
calibration stages or iterative alternation between calibration and position
reconstruction are required.
Instead, the complete detector model is optimized end-to-end within a single
differentiable learning framework.

\section{Physics-Informed Objective Function}

The proposed framework is trained exclusively through physically derived
consistency constraints.
No labelled interaction positions, calibration coefficients, or reference
trajectories are required.
Instead, supervision emerges from the requirement that the reconstructed
interaction positions and calibrated charge distributions simultaneously satisfy
the known physical relationships governing detector operation.

The objective function consists of two complementary components.
The first enforces geometric consistency between the interaction positions
reconstructed by the two MWPPAC detectors, whereas the second constrains the
calibrated charge distributions through their proportionality to the energy loss
measured in the ionization chamber.

\subsection{Geometric Consistency Loss}

For a given event, the reconstructed interaction positions
$X_1^C$, $X_2^C$, $Y_1^C$, and $Y_2^C$
must satisfy the geometric relationships imposed by the detector layout.

The predicted interaction positions are obtained using Eqs.~(\ref{eq:X2P}) and (\ref{eq:X1P}).

The geometric residuals, using Eqs.~(\ref{eq:DX1}) and (\ref{eq:DX2}), are therefore
\begin{equation}
\Delta X_1 = X_1^P - X_1^C,
\end{equation}
\begin{equation}
\Delta X_2 = X_2^P - X_2^C,
\end{equation}
\begin{equation}
\Delta Y_1 = Y_1^P - Y_1^C,
\end{equation}
\begin{equation}
\Delta Y_2 = Y_2^P - Y_2^C.
\end{equation}

The geometric consistency loss is defined as
\begin{equation}
\mathcal L_{X_{\mathrm{geom}}}
=
\frac{1}{2}
\left(
\Delta X_1^2
+
\Delta X_2^2
\right),
\end{equation}
and
\begin{equation}
\mathcal L_{Y_{\mathrm{geom}}}
=
\frac{1}{2}
\left(
\Delta Y_1^2
+
\Delta Y_2^2
\right),
\end{equation}
since the vertical and horizontal planes are independent.

Minimizing this term encourages the reconstructed interaction positions to
remain compatible with the known detector geometry and the assumption that the
particle trajectory originates from the target region.

\subsection{Charge-Energy Consistency Loss}

Position reconstruction depends only on the relative calibration of neighboring
wires.
Consequently, geometric consistency alone cannot uniquely determine the overall
normalization of the calibration coefficients.

To remove this ambiguity, an additional charge-energy consistency constraint is
introduced.
For each detector plane, the sum of calibrated wire charges is required to be
proportional to the energy deposited in one segment of the ionization chamber,
\begin{equation}
\sum_i Q_i^{\mathrm{cal}}
=
fE_{\mathrm{IC}},
\end{equation}
where $E_{\mathrm{IC}}$ denotes the ionization chamber signal and $f$ is a
fixed scaling factor.

The corresponding charge-energy consistency loss is
\begin{equation}
\mathcal L_{X_{\mathrm{charge}}}
=
\frac{1}{2}
\sum_{k}
\left(
\sum_i Q_{X_{k,i}}^{\mathrm{cal}}
-
f_{X_k} E_{\mathrm{IC}}
\right)^2,
\end{equation}
and
\begin{equation}
\mathcal L_{Y_{\mathrm{charge}}}
=
\frac{1}{2}
\sum_{k}
\left(
\sum_i Q_{Y_{k,i}}^{\mathrm{cal}}
-
f_{Y_k} E_{\mathrm{IC}}
\right)^2,
\end{equation}
where the index $k$ runs over the two detectors.

The scaling factors $f_{X_k}$ and  $f_{Y_k}$ are selected such that the resulting calibration
coefficients remain within the prescribed interval
$[0.5,2.0]$.
Since only relative wire calibration influences position reconstruction, the
precise numerical values of $f_{X_k}$ and  $f_{Y_k}$ are otherwise not physically significant.

\subsection{Joint Self-Supervised Objective}

The complete optimization objective is defined as
\begin{equation}
\mathcal L_X
=
\lambda_{X_{\mathrm{geom}}}
\mathcal L_{X_{\mathrm{geom}}}
+
\lambda_{X_{\mathrm{charge}}}
\mathcal L_{X_{\mathrm{charge}},}
\end{equation}
and 
\begin{equation}
\mathcal L_Y
=
\lambda_{Y_{\mathrm{geom}}}
\mathcal L_{Y_{\mathrm{geom}}}
+
\lambda_{Y_{\mathrm{charge}}}
\mathcal L_{Y_{\mathrm{charge}},}
\end{equation}
where
$\lambda_{\mathrm{geom}}$
and
$\lambda_{\mathrm{charge}}$
control the relative importance of the two physical constraints.

The first term constrains the reconstructed interaction positions through the
known detector geometry, whereas the second term constrains the calibration
coefficients through charge-energy consistency constraint.
Together, these complementary constraints transform detector calibration and
position reconstruction into a fully self-supervised learning problem.

The optimization therefore seeks the calibration coefficients and interaction
positions that simultaneously satisfy all available physical consistency
relations.

Importantly, neither component of the objective function requires labelled
training data.
All supervisory information is derived directly from detector physics.

\subsection{Why the Problem is Self-Supervised ?}

The proposed framework differs fundamentally from conventional supervised
learning.
No experimentally measured interaction positions are available during training,
and no reference calibration coefficients are provided.

Instead, supervision originates entirely from physically derived consistency
constraints.
The detector geometry supplies relationships between measurements obtained from
independent detector planes, while the ionization chamber provides an external
physical constraint on the total induced charge.

Consequently, the target values required for optimization are generated
implicitly from the measurement process itself rather than from human
annotations or dedicated calibration measurements.
This places the proposed method within the class of self-supervised scientific
machine learning approaches.

\section{Experimental Setup}

The proposed framework is evaluated using experimental data obtained with the VAMOS++ 
magnetic spectrometer. The objectives of the evaluation are to assess the convergence of 
the self-supervised optimization problem and quantify the quality of the reconstructed 
interaction position.

\subsection{VAMOS++ MWPPAC Detector}

The present study employs the dual position-sensitive multi-wire proportional counter 
(DPS-MWPC) configuration developed for reaction-product tracking at 
VAMOS++~(\citet{Vandebrouck2016}).
The setup comprises two low-pressure, position-sensitive MWPCs installed within a shared gas 
volume between the target and the spectrometer entrance.

The active areas of the front and back MWPCs are $40\times 61$~mm$^2$ and 
$65\times 93$~mm$^2$, respectively, aligning with the angular acceptance of 
VAMOS++. The two cathode planes are separated by $105$~mm, and the detector 
assembly is positioned $156$~mm downstream from the target. Each MWPC comprises 
a central timing cathode and two orthogonal anode planes ($X$ and $Y$) for two-dimensional position readout.

All sensing planes utilize gold-plated tungsten wires. Cathode wires have a diameter of 
$20$~$\mu$m and a pitch of $0.5$~mm; anode wires have a diameter of $20$~$\mu$m 
and a pitch of $1.0$~mm.
The cathode wires provide timing information, while the anode wires provide position information.
There are $39$, $64$, $60$, and $92$ anode wires for $X_1$, $X_2$, $Y_1$, and $Y_2$ 
planes, respectively.
The cathode-to-anode gap is 2.4~mm. The detector gas is isobutane (i-C$_4$H$_{10}$), 
typically operated at $6$~mbar pressure, with thin $0.5$~$\mu$m Mylar entrance/exit windows 
to minimize energy loss and angular straggling.

\subsection{Experimental Data}
The experimental data utilized in this study were obtained during the E850 GANIL 
experiment~\cite{DataE850}. This experiment was designed to detect, identify, 
and track light nuclei in coincidence with fission fragments produced in 
multi-nucleon transfer reactions in inverse kinematics. The reactions were induced by a 
$^{238}$U beam with an energy of $5.95$~MeV/u on a $100~\mu$g/cm$^2$ thick $^{12}$C 
target. The fission fragments were detected and isotopically identified in the VAMOS++ spectrometer. 
Light nuclei were detected in the PISTA array~\cite{BegueGuillou2026}.

\subsection{Implementation Details}

The datasets utilized for training the model consisted of  $4\times10^6$ events.
At the beginning of the training process, the dataset was randomly partitioned into the training 
set ($80~\%$) and the validation set ($20~\%$). For each epoch a random rearrangement of the
 training dataset was performed.  The training was performed in batches of
$1\times 10^3$ events. The training convergence of the neural network 
was assessed in terms of the RMSD. The duration of a single training epoch was 
approximately $6$~s. The training was stopped when RMSD($\Delta X_k$), RMSD($\Delta Y_k$)
as well as the mean values of the calibration coefficients $\overline{c_{X_1}}$,
$\overline{c_{Y_1}}$,
$\overline{c_{X_2}}$,
$\overline{c_{Y_2}}$,
reached the plateau.
Remarkably, the training convergence 
was typically achieved within approximately $0.5$~minute, indicating that the neural network swiftly and efficiently 
identifies the system characteristics. 

\begin{figure*}[]
  \centering
  \includegraphics[width=1\textwidth]{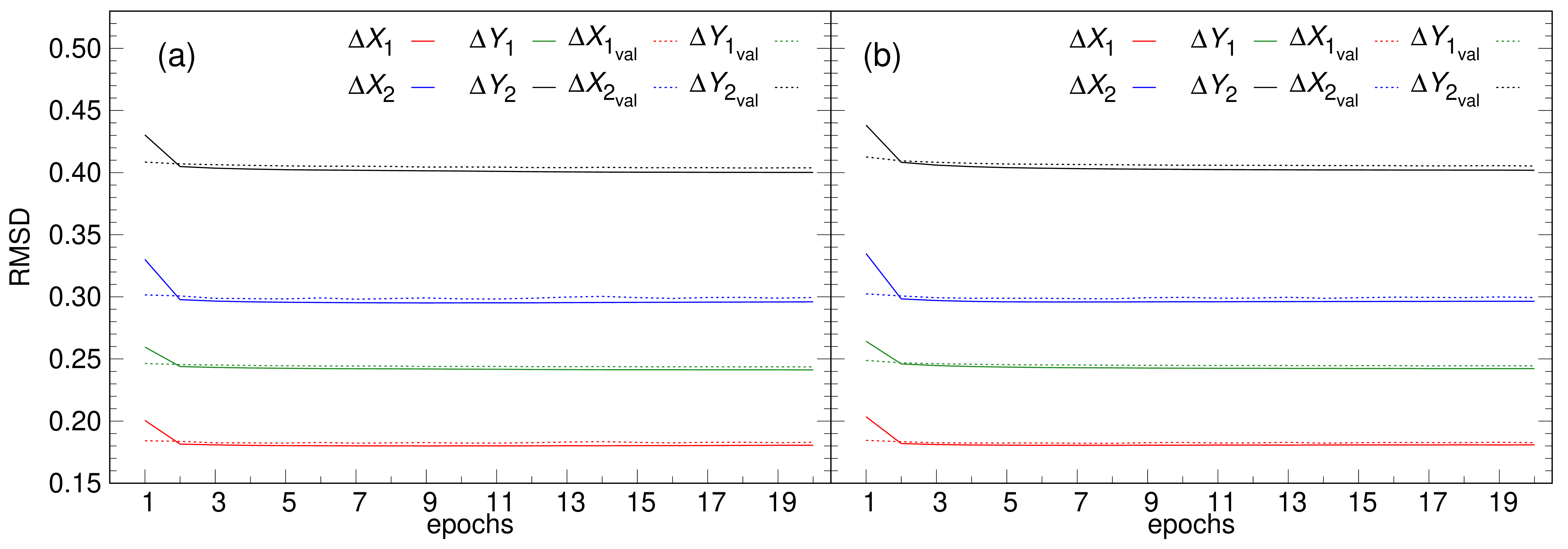}
  \caption{
    {\bf Training Convergence of Residuals:}
  The evolution of the training and validation RMSD of the residuals $\Delta X_1$, $\Delta X_2$, 
  $\Delta Y_1$, $\Delta Y_2$ obtained for the number $N$ of adjacent sensing wires, 
  see Eq.~(\ref{eq:QVEC}),  (a) $N=3$, and (b) $N=5$.
  }
  \label{fig:TrainingConvergenceResiduals}
\end{figure*}

\begin{figure*}[]
  \centering
  \includegraphics[width=1\textwidth]{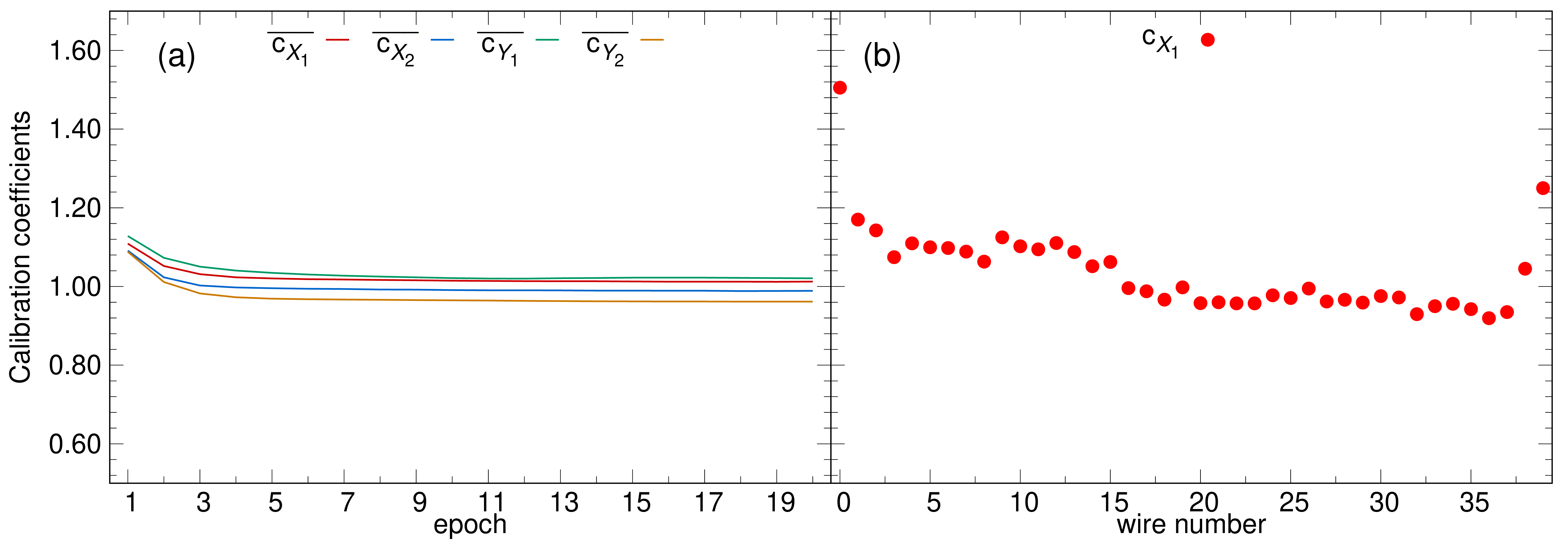}
  \caption{{\bf Training Convergence of Coefficients and Values of Coefficients:} 
  (a) The evolution of the mean value of the coefficients 
  $\overline{c_{X_1}}$, $\overline{c_{Y_1}}$,
  $\overline{c_{X_2}}$, $\overline{c_{Y_2}}$ as a function of the epoch number.
  (b) Individual values of the coefficients $c_{X_1}$ as function of wire number.
 }
  \label{fig:TrainingConvergenceCoefficients}
\end{figure*}

\begin{figure*}[]
  \centering
  \includegraphics[width=1\textwidth]{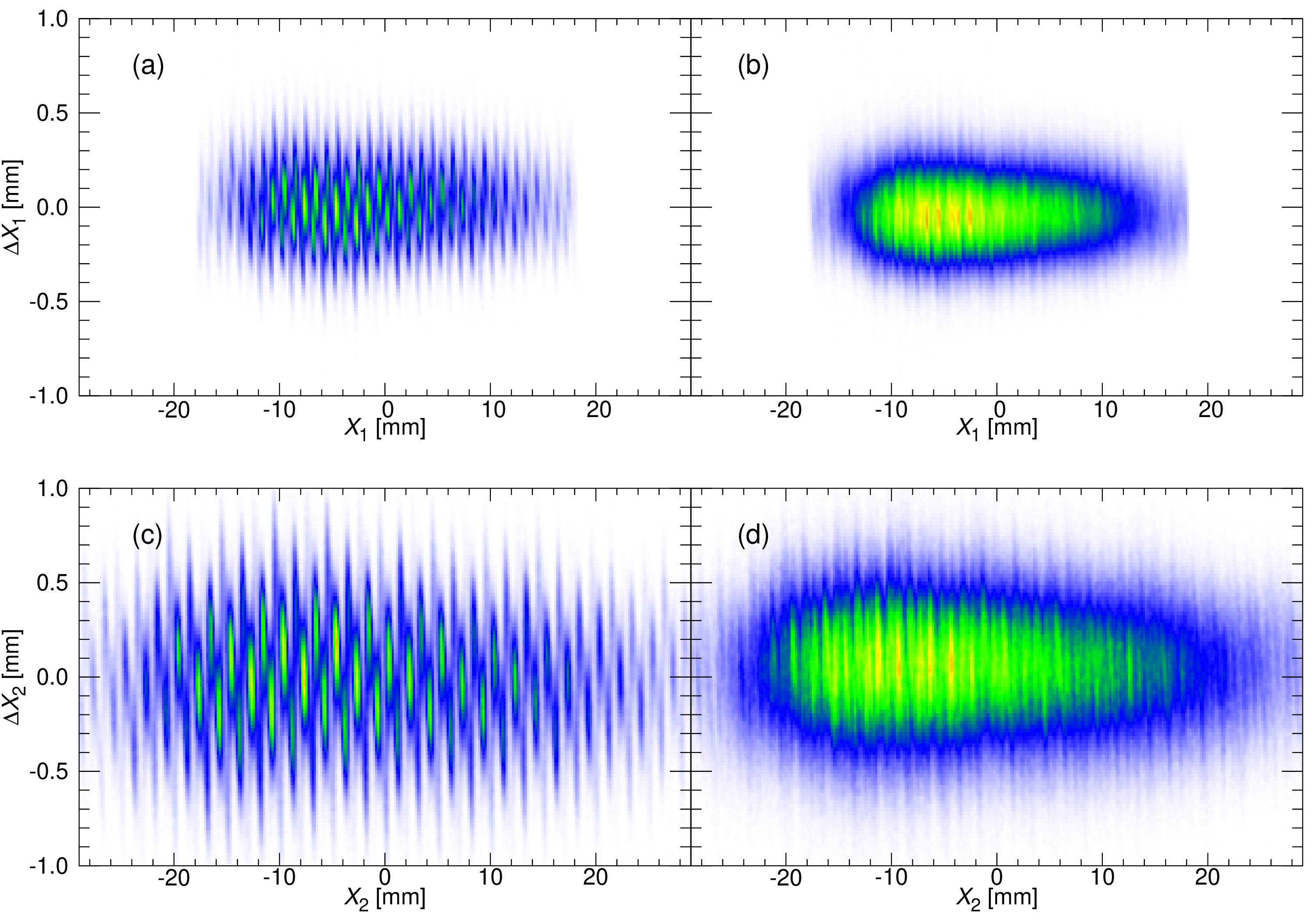}
  \caption{{\bf Spatial Distribution of Residuals:}
  Two-dimensional correlation of the residuals $\Delta X_1$ (a) and (b) and
  $\Delta X_2$ (c) and (d) as a function of the interaction position in the $X_1$
  and $X_2$ planes, respectively.
  Panels (a) and (c) illustrate the results obtained using the conventional method~(\citet{Vandebrouck2016}) 
  using hyperbolic secant function~(\citet{Lau1995}), while panels (b) and (d) illustrate the results obtained 
  in the presented framework. Both approaches used $N=3$  adjacent sensing wires.
  }
  \label{fig:SpatialDistributionResiduals}
\end{figure*}

\section{Results}

\subsection{Optimization Behavior}

The proposed self-supervised optimization is driven exclusively by physical consistency constraints. 
Consequently, the optimization converges toward detector states that simultaneously satisfy the 
geometric relationships between the MWPPAC detectors and the charge-energy consistency constraints 
imposed by the ionization chamber. Understanding the expected behavior of the optimization 
therefore provides insight into the identifiability of both the wire calibration coefficients and 
the reconstructed interaction positions.

Figure~\ref{fig:DetectorGeometry}(a) and (b) depict the simulated distributions of residuals 
$\Delta X_1$ and $\Delta X_2$ under varying conditions, assuming a true beam width of $\sigma_{X_{\rm beam}} = 0.40$~mm. 
The distributions are shown for different detector resolutions: (black) $\sigma_{X_{\rm det}} = 0$~mm, 
(red) $\sigma_{X_{\rm det}} = 0.06$~mm, and (blue) $\sigma_{X_{\rm det}} = 0.09$~mm.
The typical true beam width on the target is $\sigma_{X_{\rm beam}} = 0.40$~mm and $\sigma_{Y_{\rm beam}} = 0.50$~mm.
When the detector resolution is $\sigma_{X_{\rm det}} = 0$~mm, the widths of the residuals are $\sigma_{\Delta X_1} = 0.16$~mm and
$\sigma_{\Delta X_2} = 0.27$~mm, respectively.
As observed in the figure, since the detector resolution is much smaller than the true beam width, the widths 
of the residuals are primarily influenced by the width of the beam profile on the target.
When the detector resolution is $\sigma_{X_{\rm det}} = 0.06$~mm, the widths of the residuals are $\sigma_{\Delta X_1} = 0.18$~mm and
$\sigma_{\Delta X_2} = 0.30$~mm, respectively, while for $\sigma_{X_{\rm det}} = 0.09$~mm they are $\sigma_{\Delta X_1} = 0.19$~mm and
$\sigma_{\Delta X_2} = 0.32$~mm, respectively.
The improvement of the detector resolution by $\sim30~\%$ results in a decrease of the overall 
width of the residuals by several percent. 
Figure~\ref{fig:DetectorGeometry}(c) depicts the corresponding reconstructed distribution of the beam on the target $X_{\rm target}$. 
The anticipated widths for the detector resolution is $\sigma_{X_{\rm det}} = 0,\; 0.06,\; 0.09$~mm are
$\sigma_{X_{\rm target}} = 0.40, \; 0.44,\; 0.48$~mm, respectively. The improvement of the detector resolution by $\sim 30~\%$ results 
in a decrease of the overall width of the beam profile by $\sim 9~\%$.

\subsection{Optimization Convergence}

The first question concerns whether the proposed self-supervised optimization converges 
towards a stable detector calibration.

Figure~\ref{fig:TrainingConvergenceResiduals} depicts the evolution of the training and validation 
Root Mean Square Deviation (RMSD) of the residuals $\Delta X_1$, $\Delta X_2$, $\Delta Y_1$, 
and $\Delta Y_2$. In panel (a), the results obtained for the number of adjacent sensing wires $N=3$, 
and in panel (b), $N=5$, are shown. 

It is evident from the figure that convergence is achieved within 
a few epochs for both models. The validation curves follow closely the training curves and no overfitting
occurs. 

Notably, for the model employing $N=5$ adjacent sensing wires, 
there is no significant improvement observed as compared to the model with $N=3$.
In this study, also models utilizing $N=7,9,11$ adjacent sensing wires were tested, resulting in 
equivalent outcomes to those obtained with the model employing $N=3$ adjacent sensing wires.
Similarly, the augmentation of the complexity of the position reconstruction model did not yield 
a substantial improvement.

The RMSD values obtained for the residuals $\Delta X_1$, $\Delta X_2$, $\Delta Y_1$, 
and $\Delta Y_2$ are $0.179(1)$~mm, $0.293(1)$~mm, $0.242(1)$~mm, and $0.401(2)$~mm, respectively.
The values obtained employing $N=3$ and $N=5$ agree within $0.5~\%$
The expectation $\Delta X_1 < \Delta X_2$ and $\Delta Y_1 < \Delta Y_2$
is corroborated by the results presented in Figure~\ref{fig:TrainingConvergenceResiduals}.

Figure~\ref{fig:TrainingConvergenceCoefficients}(a) depicts the progression of the mean value of 
the coefficients $\overline{c_{X_1}}$, $\overline{c_{Y_1}}$, $\overline{c_{X_2}}$, and $\overline{c_{Y_2}}$ 
as a function of the epoch number obtained from the model employing $N=3$ adjacent sensing wires. 
It is evident from the figure that the values of the coefficients 
jointly stabilize as the optimization process converges. Figure~\ref{fig:TrainingConvergenceCoefficients}(a) 
illustrate the corresponding $39$ individual converged coefficients for the $X_1$ plane. 

\subsection{Spatial distribution of Residuals}

Figure~\ref{fig:SpatialDistributionResiduals} illustrates the two-dimensional correlation 
of the residuals $\Delta X_1$ and $\Delta X_2$ as a function of the interaction position 
in the $X_1$ and $X_2$ planes, respectively. The results obtained using the conventional 
method~(\citet{Vandebrouck2016}) employing the hyperbolic secant function~(\citet{Lau1995}) 
are presented in Figure~\ref{fig:SpatialDistributionResiduals}(a) and (c), while the results obtained 
using the proposed framework are depicted in Figure~\ref{fig:SpatialDistributionResiduals}(b) and (d). 
Both approaches utilized $N=3$ adjacent sensing wires.

The conventional method's results exhibit pronounced local discontinuities, with the interaction 
positions clustering near the wires because the assumed analytical induction profile cannot perfectly 
represent the true detector response.
In contrast, the proposed framework removes these artifacts because the induction profile is learned implicitly during 
optimization rather than prescribed a priori.
The framework therefore not only improves the global width of the residuals but also handles
the local discontinuities related to the form of the induction profile. The widths obtained using the presented framework 
are summarized and compared to those obtained by the conventional method in Table~\ref{WidthSummary}. 
A mean improvement of the widths of  $\delta \sigma =0.017$~mm can be deduced from the table.

\begin{table}
\caption{
Summary of the widths of the residuals, 
the reconstructed position on the target and the position resolution, inferred using the reference wires,
obtained employing the conventional method and the presented framework.
\label{WidthSummary}}
\renewcommand{\arraystretch}{1.3} % 30% taller rows

\centering
\vspace{0.3em}
\begin{tabular*}{0.95\columnwidth}{@{\extracolsep{\fill}}ccc}
\hline
 &  this work (mm) & Conventional (mm) \\

\hline
$\sigma_{\Delta X_1}$          & 0.179(1) & 0.193(1) \\
$\sigma_{\Delta X_2}$          & 0.293(1) & 0.317(1) \\
$\sigma_{\Delta Y_1}$          & 0.242(1) & 0.252(1) \\
$\sigma_{\Delta Y_2}$          & 0.401(2) & 0.418(1) \\
\hline
$\sigma_{X_{\rm target}}$ & 0.433(2) & 0.468(2) \\
$\sigma_{Y_{\rm target}}$ & 0.572(2) & 0.592(2) \\
\hline
$\sigma_{W_{1}}$ & 0.06(1) & 0.09(2) \\
$\sigma_{W_{2}}$ & 0.05(1) & 0.09(2) \\

\end{tabular*}
\end{table}

\begin{figure*}[htbp]
  \includegraphics[width=0.95\textwidth]{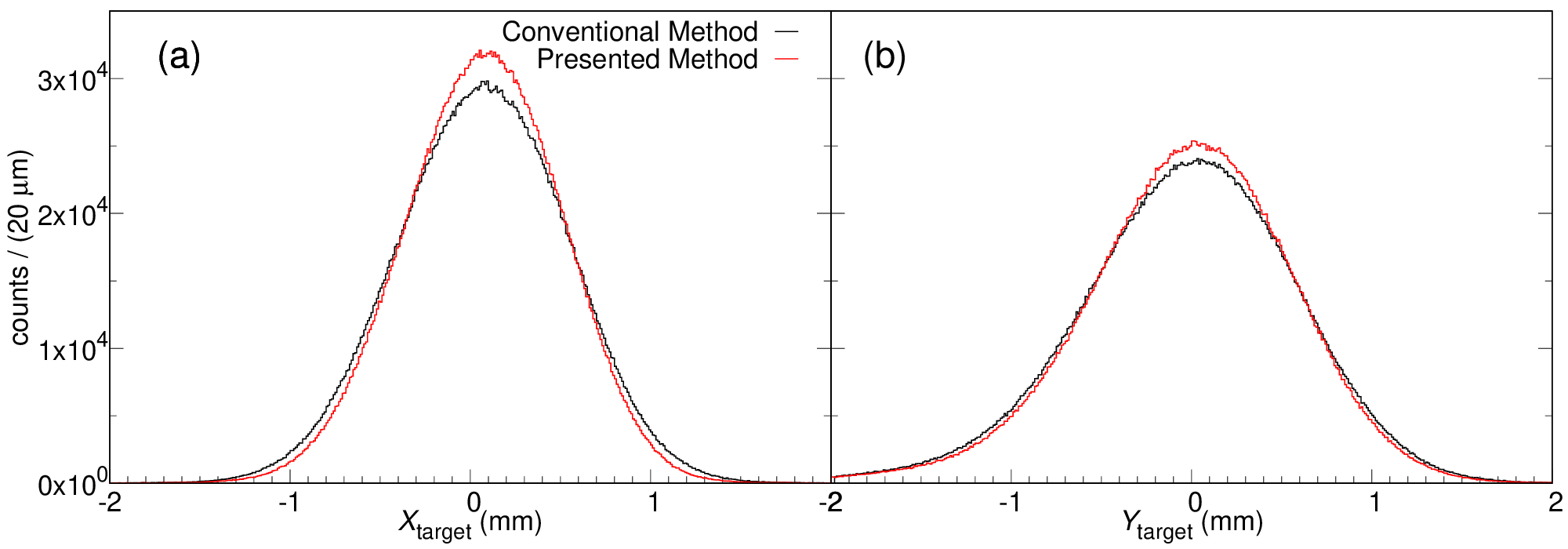}
  \caption{{\bf Reconstructed Beam Profile:}   Reconstructed beam profile on the target (a) $X_{\rm target}$ and (b) $Y_{\rm target}$.
  The results obtained using the conventional method~(\citet{Vandebrouck2016})   using hyperbolic secant 
  function~(\citet{Lau1995}) are indicated in black, while the results obtained    in the presented framework are indicated in red. 
  Both approaches used $N=3$  adjacent sensing wires.  \label{fig:ReconstructedBeamProfile}}
\end{figure*}

\subsection{Reconstructed Beam Profile on the Target}

Figure~\ref{fig:ReconstructedBeamProfile} depicts the reconstructed beam profile on the target, 
specifically in the $X_{\rm target}$ and $Y_{\rm target}$ planes. The results obtained using the 
conventional method are represented in black, while the results obtained using the presented 
framework are depicted in red. A notable improvement is evident in the figure when comparing 
the results obtained using the presented framework to those obtained using the conventional method
since a systematic reconstruction bias has been removed.
The resulting widths for the conventional approach are $\sigma_{X_{\rm target}} = 0.468(2)$~mm 
and $\sigma_{Y_{\rm target}} = 0.592(2)$~mm, whereas the resulting widths for the presented framework 
are $\sigma_{X_{\rm target}} = 0.433(2)$~mm and $\sigma_{Y_{\rm target}} = 0.572(2)$~mm.

\subsection{Position Resolution}

\begin{figure*}[htbp]
  \centering
  \includegraphics[width=0.95\textwidth]{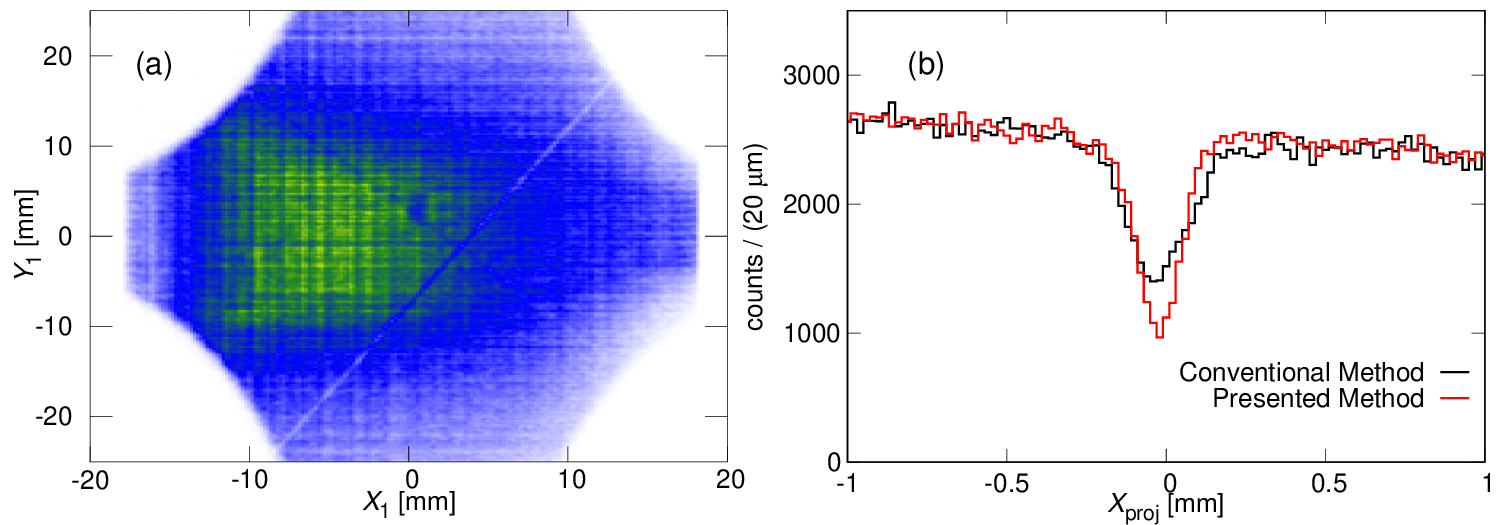}
  \caption{{\bf Position Resolution}
  (a) Two-dimensional projection of the events on the plane in front of the detector, where 
  the $100~\mu$m reference wire is located. The projection was obtained using the current framework.
  (b) One-dimensional projection on a plane perpendicular to the reference wire.
  The results obtained the conventional method are indicated in black and the results obtained 
  the proposed framework are indicated in red.
  }
  \label{fig:ReferenceWire}
\end{figure*}

Two gold-plated tungsten wires with a diameter of $100~\mu$m are positioned diagonally 
$10$~mm upstream and downstream of the front and back MWPC, respectively.
Figure~\ref{fig:ReferenceWire}(a) depicts the two-dimensional projection of the events onto the 
plane in front of the detector, where the $100~\mu$m reference wire is located. The shadow 
of the diagonally placed front reference wire is distinctly visible. Despite the events being 
projected onto the plane in front of the detector, the perpendicularly placed back reference 
wire can also be discerned.

Figure~\ref{fig:ReferenceWire}(b) depicts the one-dimensional projection on a plane perpendicular 
to the reference wire. The improvement in the reference wire width definition using the presented
framework, as compared to the conventional method, is clearly visible in the figure.
This spectrum can be used to obtain the position resolution~(\citet{Vandebrouck2016}),
assuming the shadowing of a wire with $100~\mu$m diameter
leading to $\sigma_{W_1} = 0.09(2)$~mm for the conventional method and $\sigma_{W_1} = 0.06(1)$~mm
for the presented method.
The same procedure was applied for the reference wire placed at the back of the detector 
leading to $\sigma_{W_2} = 0.09(2)$~mm for the conventional method and $\sigma_{W_2} = 0.05(1)$~mm.
The position resolution achieved using the presented framework demonstrates a substantial 
enhancement of $33~\%$ and $44~\%$ for front and rear reference wires, respectively.

Unlike the previous results presented above, which are derived from the self-supervised optimization itself, 
the reference-wire measurement constitutes an entirely independent validation of the reconstructed positions.

\subsection{Implicit Learning of the Induction Profile}

Conventional interaction position reconstruction algorithms rely on an explicit
analytical model describing the charge induced on neighboring sensing wires.
Typical assumptions include Gaussian, hyperbolic secant, or related parametric
profiles whose parameters are determined through analytical fitting.
The accuracy of the reconstructed interaction position therefore depends
directly on the validity of the assumed induction model.

The proposed framework follows a fundamentally different strategy.
No analytical expression describing the induction profile is introduced during
training.
Instead, the shared position reconstruction network learns the mapping between
local calibrated charge distributions and sub-wire interaction positions solely
through the physics-informed self-supervised objective.

The absence of local discontinuities in
Fig.~\ref{fig:SpatialDistributionResiduals}
provides strong evidence that the learned representation accurately captures the
true detector response.
Rather than approximating the induction profile using a predefined functional
form, the network implicitly reconstructs the detector response directly from
experimental observations.
This considerably reduces modeling bias while simultaneously improving the
uniformity of the reconstructed interaction positions across the detector
surface.

\subsection{Transferability and Generalization}

An important property of the proposed framework is that the position
reconstruction network operates exclusively on local calibrated charge
distributions and therefore remains independent of the detector geometry,
absolute wire numbering, and active detector area.

Consequently, the same reconstruction model can be applied to MWPPAC detectors
with different dimensions without modification of the network architecture.
Only the detector-specific calibration subnetworks need to be re-optimized,
whereas the learned local relationship between neighboring wire charges and
sub-wire interaction position remains unchanged.

This property is particularly relevant for the VAMOS++ focal-plane tracking
detectors, which employ substantially larger active areas than the entrance
MWPPAC system investigated in the present work.
The proposed formulation therefore provides a generic reconstruction framework
whose applicability extends beyond the detector configuration considered here.

\section{Discussion}

The results presented in this work demonstrate that accurate wire calibration
and interaction position reconstruction can be achieved without labelled
training data or dedicated detector calibration measurements.
Rather than learning from externally provided target values, the proposed
framework exploits the intrinsic redundancy of the detector system together
with known physical relationships governing particle transport and detector
response.
This transforms detector calibration from a supervised parameter estimation
problem into a self-supervised optimization problem in which the supervisory
signal is generated directly by the measurement process itself.

A central aspect of the proposed formulation is the simultaneous optimization
of two fundamentally different categories of latent variables.
The wire calibration coefficients represent global detector parameters that are
shared by all recorded events, whereas the interaction positions correspond to
local event-specific variables.
Although these quantities are conceptually distinct, they become strongly
coupled through the physical consistency constraints introduced in the
optimization objective.
Improved wire calibration produces more accurate local charge distributions,
which facilitate more precise position reconstruction.
Conversely, improved interaction positions increase the geometric consistency
between the two MWPPAC detectors, thereby providing a stronger supervisory
signal for calibration.
The two optimization processes therefore reinforce each other throughout
training until a physically consistent solution is reached.

Another important characteristic of the proposed framework is its robustness to
imperfect detector signals.
Since the position reconstruction network operates on the complete local charge
distribution rather than relying on a single measurement, the interaction
position is inferred from the collective response of neighboring sensing wires.
Consequently, the reconstruction remains possible even when individual charge
measurements become unreliable or unavailable.
This property is particularly relevant for experiments involving a wide range
of ion species, where electronic gain settings are frequently optimized to
preserve detection efficiency for light ions, leading to saturation of the
maximum induced charge for heavier reaction products.
Preliminary investigations indicate that the proposed framework could be 
used to learn to reconstruct interaction positions accurately even after removing 
the saturated maximum charge from the network input, demonstrating that the learned
representation effectively exploits the redundancy of the neighboring wire
signals.
Although a detailed analysis of this capability is beyond the scope of the
present work, it illustrates the potential robustness of learned
physics-informed representations under realistic experimental conditions.

Beyond the specific application to the entrance MWPPAC detectors, the proposed
framework illustrates a more general principle for intelligent scientific
instrumentation.
The neural network architecture itself is largely independent of the detector
under consideration.
Instead, detector-specific knowledge is introduced exclusively through the
physics-informed objective function.
For the entrance MWPPAC system, supervision originates from geometric
consistency between two independent position measurements.
For the ionization chamber investigated in our previous work~(\citet{Rejmund2026arxiv}), 
supervision is provided by ionic charge-state and atomic mass number consistency.
Similarly, for the focal-plane MWPPAC detectors of VAMOS++, equivalent
self-supervised calibration can be achieved by exploiting the dependence of the
reconstructed atomic mass number on the measured interaction positions.
Consequently, the proposed methodology should be viewed not as an isolated
calibration algorithm, but rather as a modular framework in which detector
calibration is obtained by combining differentiable detector models with the
appropriate physical consistency constraints.

Taken together with our previous work on self-supervised ionization chamber
calibration, the present results suggest a broader perspective for the
development of autonomous scientific instrumentation.
Within the VAMOS++ spectrometer, only the timing calibration currently requires
a conventional calibration procedure.
Trajectory reconstruction is already performed in an absolute coordinate system,
while both the ionization chamber and the MWPPAC detectors can now be calibrated
through self-supervised optimization.
The remaining focal-plane MWPPAC calibration naturally follows the same
methodology using detector-specific physical constraints.
This progression indicates that complete self-supervised calibration of the
spectrometer is becoming a realistic objective.

More generally, the proposed framework demonstrates that modern scientific
instruments can increasingly exploit their own physical redundancy to perform
continuous self-calibration during normal operation.
Rather than relying on dedicated calibration measurements performed before an
experiment, future intelligent detector systems may continuously adapt their
internal calibration parameters as detector characteristics evolve due to
changing operating conditions, electronic drifts, or aging.
Such adaptive behavior has the potential to improve measurement reliability,
reduce experimental downtime, and simplify the operation of increasingly
complex scientific facilities.

The proposed framework nevertheless has several limitations.
Its success depends on the availability of sufficiently informative physical
constraints capable of uniquely determining the latent calibration parameters.
If the detector geometry does not provide adequate redundancy or if the
underlying physical model is incomplete, additional supervisory information may
be required.
Furthermore, although the present work demonstrates the methodology using
linear gain calibration, the extension to more complex nonlinear detector
responses will require richer parameterizations and potentially stronger
regularization strategies.
These directions constitute promising avenues for future research.

\section{Conclusions}

This work presented a physics-informed self-supervised framework for the joint calibration of 
sensing-wire gains and interaction position reconstruction in Multi-Wire Parallel Plate Avalanche Counters. 
Unlike conventional calibration procedures, the proposed method requires neither dedicated calibration measurements 
nor labelled interaction positions. Instead, calibration and reconstruction are formulated as a unified latent optimization 
problem in which supervision is derived exclusively from detector geometry and charge-energy consistency constraints.

The proposed framework jointly estimates global wire calibration coefficients and event-wise interaction positions 
through end-to-end differentiable optimization. A detector-independent position reconstruction network learns the 
relationship between local charge distributions and sub-wire interaction positions directly from experimental data, 
eliminating the need to assume analytical induction profiles. Experimental results obtained with the entrance MWPPAC 
tracking detectors of the VAMOS++ magnetic spectrometer demonstrate stable optimization, improved spatial homogeneity, 
and enhanced position reconstruction performance while enabling detector calibration to be performed continuously during 
normal operation.

Beyond the specific application considered here, the proposed methodology illustrates a general paradigm for intelligent 
scientific instrumentation. Rather than designing detector-specific calibration algorithms, calibration is formulated as a 
modular self-supervised learning problem in which detector-specific physical knowledge enters only through differentiable 
consistency constraints. This principle is applicable to a broad range of detector systems wherever redundant measurements 
or well-established physical relationships provide sufficient supervisory information.

Together with our previous work on self-supervised calibration of the VAMOS++ ionization chamber, the present study 
substantially extends the range of detector subsystems that can be calibrated autonomously. The remaining focal-plane 
MWPPAC detectors naturally fit within the same framework by replacing geometric consistency with atomic mass-reconstruction 
consistency as the supervisory constraint. 

Consequently, the proposed methodology represents an important step toward 
fully self-supervised scientific instrumentation, capable of continuously adapting its calibration to detector aging, 
electronic drifts, and changing experimental conditions without interrupting normal operation. In this paradigm, the detector 
is no longer calibrated before the experiment; it is calibrated by the experiment itself.

Future work will investigate the extension of the framework to detector responses affected by signal saturation, 
nonlinear calibration models, and additional detector subsystems. More broadly, we believe that combining differentiable 
machine learning models with detector-specific physical constraints provides a promising foundation for the next 
generation of adaptive, autonomous scientific instruments.

\bibliography{references}

@misc{DataE850,
author = {{E850-21 Collaboration}},
title = {E850-21 {GANIL} Dataset},
doi = {10.26143/ganil-2023-e850_21},
year = {2024}
}

@article{BegueGuillou2026,
author = {B{\'{e}}gu{\'{e}}–Guillou, L and Lemasson, A and Morfouace, P and Ramos, D and Taieb, J and Frankland, J D and Rejmund, M and Fremont, G and Gangnant, P and Cobo-Zarzuelo, A and Kumar, N and Efremov, T and Chatillon, A and Cl{\'{e}}ment, E and {De France}, G and Francheteau, A and Jangid, I and Lenain, C and Mauss, B and Tanaka, T and Audoin, L and Caamano, M and Errandonea, B and Godio, M and Gruyer, D and Jacquot, B and Lalande, M and Malone, R C and Munoz, A and Ramirez, A P D and Rodr{\'{i}}guez-S{\'{a}}nchez, J L and Schmitt, C and Syrett, O and Surrault, C and Tonchev, A P},
doi = {https://doi.org/10.1016/j.nima.2026.171671},
issn = {0168-9002},
journal = {Nuclear Instruments and Methods in Physics Research Section A},
pages = {171671},
title = {{Performance of the Particle-Identification Silicon-Telescope Array coupled with the VAMOS++ magnetic spectrometer}},
url = {https://www.sciencedirect.com/science/article/pii/S0168900226003979},
volume = {1090},
year = {2026}
}

@article{Pullanhiotan2008,
author = {Pullanhiotan, S. and Rejmund, M. and Navin, A. and Mittig, W. and Bhattacharyya, S.},
doi = {10.1016/j.nima.2008.05.003},
issn = {01689002},
journal = {Nuclear Instruments and Methods in Physics Research Section A},
month = {aug},
number = {3},
pages = {343--352},
title = {{Performance of VAMOS for reactions near the Coulomb barrier}},
url = {http://www.sciencedirect.com/science/article/pii/S0168900208007080},
volume = {593},
year = {2008}
}

@article{Rejmund2011,
author = {Rejmund, M. and Lecornu, B. and Navin, A. and Schmitt, C. and Damoy, S. and Delaune, O. and Enguerrand, J.M. and Fremont, G. and Gangnant, P. and Gaudefroy, L. and Jacquot, B. and Pancin, J. and Pullanhiotan, S. and Spitaels, C.},
doi = {10.1016/j.nima.2011.05.007},
issn = {01689002},
journal = {Nuclear Instruments and Methods in Physics Research Section A},
month = {aug},
number = {1},
pages = {184--191},
title = {{Performance of the improved larger acceptance spectrometer: VAMOS++}},
url = {http://www.sciencedirect.com/science/article/pii/S0168900211008515},
volume = {646},
year = {2011}
}

@article{Rejmund2025a,
author = {Rejmund, M and Lemasson, A},
doi = {10.1016/j.nima.2025.170445},
issn = {01689002},
journal = {Nuclear Instruments and Methods in Physics Research Section A: Accelerators, Spectrometers, Detectors and Associated Equipment},
month = {jul},
pages = {170445},
title = {{Seven-dimensional trajectory reconstruction for VAMOS++}},
url = {https://www.sciencedirect.com/science/article/pii/S0168900225002463 https://linkinghub.elsevier.com/retrieve/pii/S0168900225002463},
volume = {1076},
year = {2025}
}

@article{Rejmund2025b,
author = {Rejmund, M and Lemasson, A},
doi = {10.1088/1748-0221/20/08/P08022},
journal = {Journal of Instrumentation},
month = {aug},
number = {08},
pages = {P08022},
publisher = {IOP Publishing},
title = {{Analysis of atomic charge state and atomic number for VAMOS++ magnetic spectrometer using deep neural networks and fractionally labelled events}},
url = {https://dx.doi.org/10.1088/1748-0221/20/08/P08022},
volume = {20},
year = {2025}
}

@misc{Rejmund2026arxiv,
      title={Self-Supervised Calibration of Scientific Instruments Using Physical Consistency Constraints}, 
      author={M. Rejmund and A. Lemasson},
      year={2026},
      eprint={2606.29466},
      archivePrefix={arXiv},
      primaryClass={cs.LG},
      url={https://arxiv.org/abs/2606.29466}, 
}

@article{Lau1995,
title = {Optimization of centroid-finding algorithms for cathode strip chambers},
journal = {Nuclear Instruments and Methods in Physics Research Section A: Accelerators, Spectrometers, Detectors and Associated Equipment},
volume = {366},
number = {2},
pages = {298-309},
year = {1995},
issn = {0168-9002},
doi = {https://doi.org/10.1016/0168-9002(95)00604-4},
url = {https://www.sciencedirect.com/science/article/pii/0168900295006044},
author = {Kwong Lau and Jörg Pyrlik}
}

@article{Vandebrouck2016,
author = {Vandebrouck, M. and Lemasson, A. and Rejmund, M. and Fremont, G. and Pancin, J. and Navin, A. and Michelagnoli, C. and Goupil, J. and Spitaels, C. and Jacquot, B.},
doi = {10.1016/j.nima.2015.12.040},
issn = {01689002},
journal = {Nuclear Instruments and Methods in Physics Research Section A},
month = {mar},
pages = {112--117},
title = {{Dual Position Sensitive MWPC for tracking reaction products at VAMOS++}},
url = {http://www.sciencedirect.com/science/article/pii/S0168900215016113},
volume = {812},
year = {2016}
}

\end{document}